\def\year{2020}\relax
\documentclass[letterpaper]{article} 
\usepackage{aaai20}  
\usepackage{times}  
\usepackage{helvet} 
\usepackage{courier}  
\usepackage[hyphens]{url}  
\usepackage{graphicx} 
\usepackage{graphicx}  
\newcommand{\etal}{\textit{et al}.~}

\newcommand{\ieno}{\textit{i}.\textit{e}.}

\newcommand{\egno}{\textit{e}.\textit{g}.} 

\usepackage{algorithm}
\usepackage{algorithmicx}
\usepackage{algpseudocode}
\usepackage{amsmath}
\usepackage{booktabs}       
\usepackage{multirow}

\usepackage{enumitem}
\usepackage{xcolor}
\usepackage{amsfonts}

\title{Uncertainty-Aware Multi-Shot Knowledge Distillation \\ for Image-Based Object Re-Identification}

\author{
Xin Jin$^{1}$\thanks{This work was done when Xin Jin was an intern at MSRA.} \qquad Cuiling Lan$^{2}$\thanks{Corresponding Author.} \qquad Wenjun Zeng$^{2}$ \qquad Zhibo Chen$^{1\dag}$ \\
University of Science and Technology of China$^{1}$ \qquad Microsoft Research Asia$^{2}$ \\
{\tt\small jinxustc@mail.ustc.edu.cn\quad \{culan, wezeng\}@microsoft.com\quad chenzhibo@ustc.edu.cn}}

\begin{document}

\maketitle

\begin{abstract}
Object re-identification (re-id) aims to identify a specific object across times or camera views, with the person re-id and vehicle re-id as the most widely studied applications. Re-id is challenging because of the variations in viewpoints, (human) poses, and occlusions. Multi-shots of the same object can cover diverse viewpoints/poses and thus provide more comprehensive information. In this paper, we propose exploiting the multi-shots of the same identity to guide the feature learning of each individual image. Specifically, we design an Uncertainty-aware Multi-shot Teacher-Student (UMTS) Network. It consists of a teacher network (T-net) that learns the comprehensive features from multiple images of the same object, and a student network (S-net) that takes a single image as input. In particular, we take into account the data dependent \emph{heteroscedastic uncertainty} for effectively transferring the knowledge from the T-net to S-net. To the best of our knowledge, we are the first to make use of multi-shots of an object in a teacher-student learning manner for effectively boosting the single image based re-id. We validate the effectiveness of our approach on the popular vehicle re-id and person re-id datasets. In inference, the S-net alone significantly outperforms the baselines and achieves the state-of-the-art performance.
\end{abstract}

\section{Introduction}

Object re-identification (re-id) aims to identify/match a specific object in different places, times, or camera views, from either images or video clips, for the purpose of tracking or retrieval. Because of the high demand in practice, person re-id and vehicle re-id are two dominant research areas for object re-id. In this work, we focus on the popular image-based person and vehicle re-id tasks.

\begin{figure}
  \centerline{\includegraphics[width=1.0\linewidth]{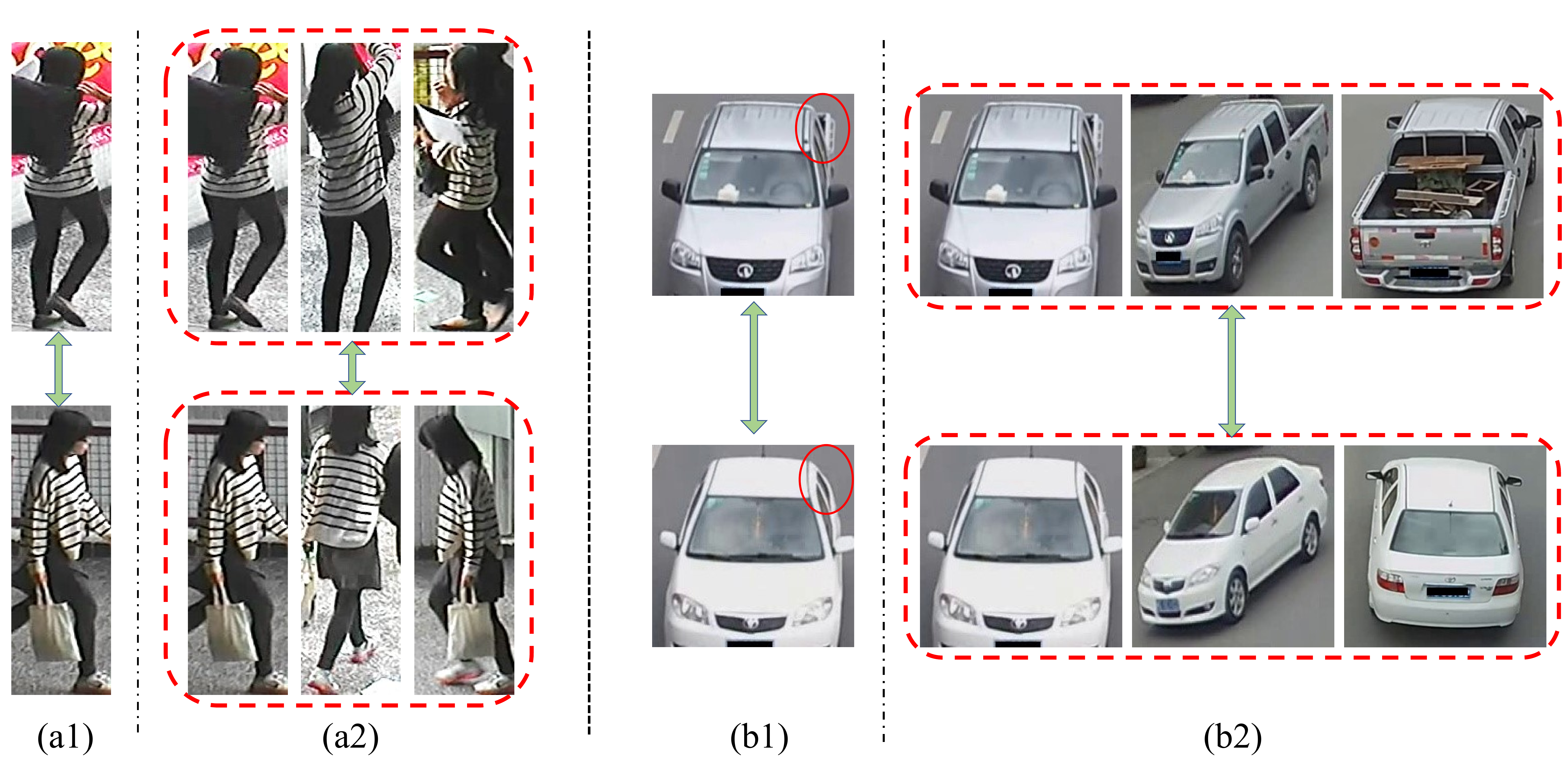}}
  \caption{Challenges {in image-based re-id}: (a1) Inconsistency of visible body regions; and (b1) Lack of comprehensive information from a single image. Observation and Motivation: multi-shot images make it easier to identify whether they are the same person/vehicle {as shown in (a2) and (b2)}.}
  \centering
\label{fig:motivation}
\end{figure}

Images to be matched typically have large variations in terms of capturing viewpoints, (human) poses, lighting, and occlusions, making re-id a challenging task \cite{subramaniam2016deep,su2017pose,li2017learning,zhao2017spindle,ge2018fd,qian2018pose,zhang2019DSA,wang2017orientation,RAM2018}. These result in inconsistency of visible object regions across images (see Fig. \ref{fig:motivation}(a1)) and lack of comprehensive information from a single image (see Fig. \ref{fig:motivation}(b1)) for matching, and spatial semantics misalignment. 

\begin{figure*}[th]
  \centerline{\includegraphics[width=1.0\linewidth]{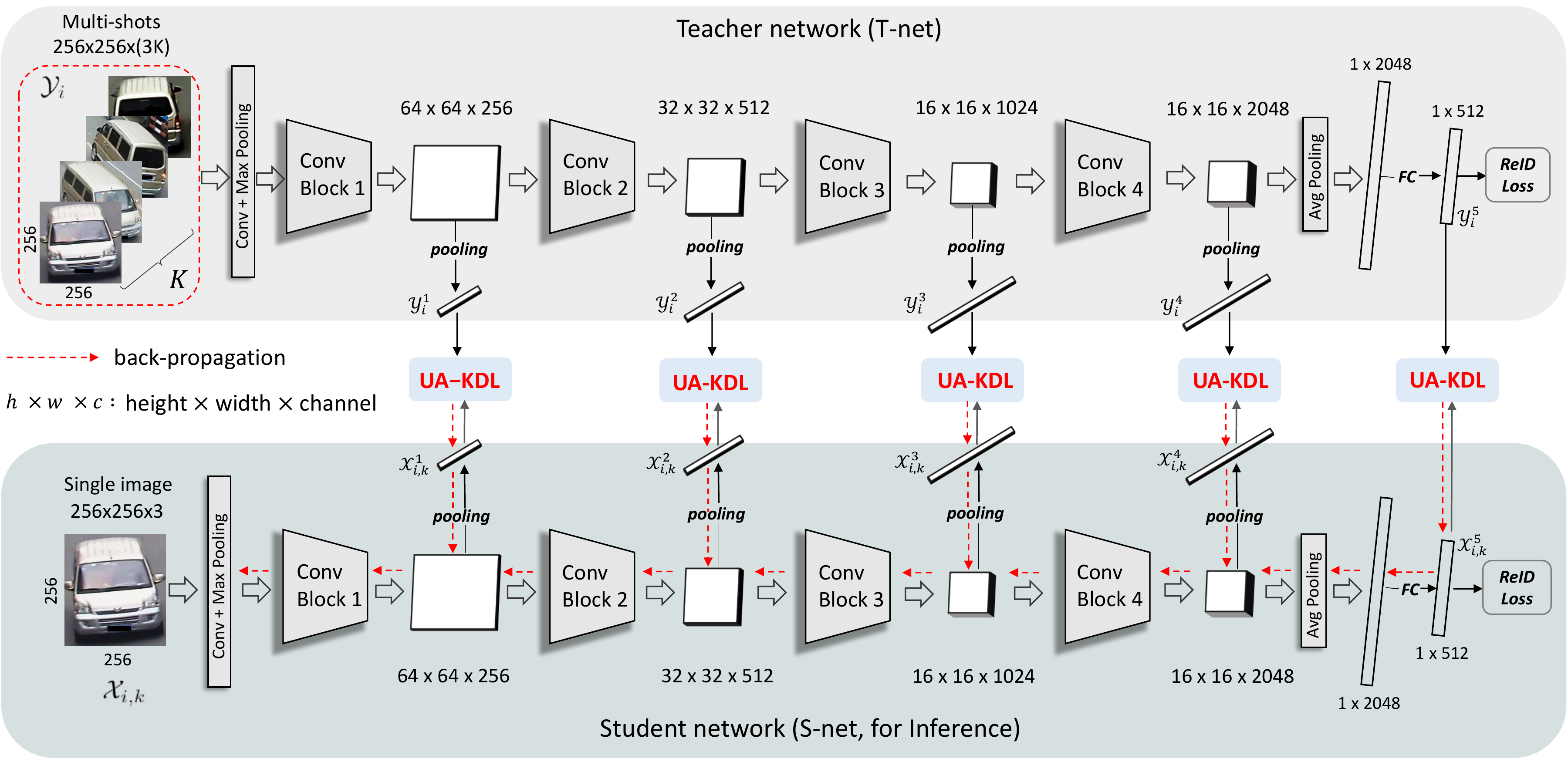}}
  \caption{Proposed Uncertainty-aware Multi-shot Teacher-Student (UMTS) Network. It consists of a Teacher-network (T-net) that learns features from the concatenation of multi-shots (\ieno, $K$ images, $\mathcal{Y}_{i}$) of the same identity $i$, and a Student-network (S-net) that takes a single image $\mathcal{X}_{i,k}$ of the $K$ images as input. To enable efficient feature learning from the T-net, we take into account the data dependent \emph{heteroscedastic uncertainty} and design an Uncertainty-aware Knowledge Distillation Loss (UA-KDL), which we apply at different layers/stages of the teacher-student network. During inference, we use only the S-net.} 
  \centering
\label{fig:framework}
\end{figure*}

Generally, for a specific object (\egno, person, vehicle), multiple images captured from different viewpoints or times can provide more comprehensive information, making the identification much easier (see Fig. \ref{fig:motivation}(a2) and (b2)). For example, the difference between the  rear of the vehicle is difficult to identify from the two single images in Fig. \ref{fig:motivation}(b1) but the difference becomes very obvious when comparing the two sets of multi-shot images shown in Fig. \ref{fig:motivation}(b2). It is worth noting that for image-based re-id, only a single image is available as a query during  inference/testing. The exploration of comprehensive information of multi-shot images is underexplored and remains an open problem.

In this paper, we propose an Uncertainty-aware Multi-shot Teacher-Student (UMTS) Network for exploiting the multi-shot images to enhance the image-based object re-id performance in a teacher-student manner, without increasing the inference complexity or changing the inference setting. We achieve this by distilling knowledge from the multi-shots of the same object and applying it to guide single shot network learning. Fig. \ref{fig:framework} shows the flowchart of the proposed Uncertainty-aware Multi-shot Teacher-Student (UMTS) Network. It consists of a Teacher network (T-net) that learns features from the multi-shots (\ieno, $K$ images) of the same object, and a Student network (S-net) that takes a single image of the $K$ images as input. In particular, different individual images from the multi-shots have different visible object regions, occlusions, and image quality, and thus different capabilities in approaching the knowledge of the multi-shot images. We take into account the data dependent \emph{heteroscedastic uncertainty} \cite{kendall2017uncertainties} and design an Uncertainty-aware Knowledge Distillation Loss (UA-KDL) to enable efficient learning of the S-net from the T-net. We conduct extensive ablation studies and demonstrate the effectiveness of the framework and components on both person re-id and vehicle re-id datasets. Our main contributions are summarized as follows:

\begin{itemize}
\item We propose a powerful Uncertainty-aware Multi-shot Teacher-Student (UMTS) Network to exploit the comprehensive information of multi-shots of the same object for effective single image re-id, without increasing computational cost in inference. 
\item We take into account the data dependent \emph{heteroscedastic uncertainty} and design an Uncertainty-aware Knowledge Distillation Loss (UA-KDL), which can efficiently regularize the feature learning at different semantics levels (\ieno, layers/stages).
\item To the best of our knowledge, we are the first to make use of multi-shots of an object in a teacher-student learning manner for efficient image-based re-id. 
\end{itemize}

\section{Related Work}

\subsection{Image-based Person/Vehicle Re-ID}

For image-based person re-id, a lot of efforts are made to address spatial semantics misalignment problem, \ieno, across images the same spatial positions usually do not correspond to the same body parts. Many approaches tend to make explicit use of semantic cues such as pose (skeleton), to align the body parts \cite{kalayeh2018human,suh2018part,liu2018pose,qian2018pose,ge2018fd,zhang2019DSA}. Some approaches leverage attention designs to selectively focus on different body parts of the person \cite{liu2017end,zhao2017deeply}. Some other approaches split the feature map to rigid grids for the coarse alignment and jointly consider the global feature and local details \cite{sun2017beyond,wang2018learning}. Moreover, several works use GAN to augment the training data with pseudo labels assigned to remedy the insufficiency of training samples \cite{zheng2017unlabeled,huang2018multi,zheng2019joint}. To address the viewpoint variation problem for vehicle re-id, Zhou \etal design a conditional generative network to infer cross-view images and then combine the features of the input and generated views to improve the re-id \cite{zhou2017cross}. In \cite{zhou2018aware}, a complex multi-view feature inference scheme is proposed based on an attention and GAN based model.

Different from the above works, we aim to explore the comprehensive information of multi-shot images of an object in a teacher-student manner to improve single image based re-id. It is a general re-id {framework} and we validate its effectiveness for both person re-id and vehicle re-id.

\subsection{Teacher-Student Learning}
Recent studies show that the knowledge learned by a strong teacher network can improve the performance of a student network \cite{chen2017learning,zhou2018rocket,wang2019progressive}. Hinton \etal propose distilling the knowledge in an ensemble of models into a single model \cite{hinton2015distilling}. Romero \etal extend this idea to enable the training of a student that is deeper and thinner than the teacher using both the outputs and the intermediate representations learned by the teacher \cite{romero2015fitnets}. Most existing methods focus on learning a light-weight student model from a teacher with the same input data. In contrast, our work aims to distill knowledge from multi-shot images to teach a single shot image feature learning for robust re-id.

\subsection{Uncertainty and Heteroscedastic Uncertainty}
In Bayesian viewpoint, there are two main types of uncertainty: \emph{epistemic} uncertainty and \emph{aleatoric} uncertainty  \cite{kendall2017uncertainties,gal2016uncertainty}. \emph{Epistemic} uncertainty accounts for uncertainty in the model parameters, which is often referred to as \emph{model uncertainty}. \emph{Aleatoric} uncertainty can further be categorized into \emph{homoscedastic uncertainty}, which stays constant for different input data and varies between different tasks, and \textbf{\emph{heteroscedastic uncertainty}}, which depends on the inputs to the model, with some noisy inputs potentially having poorer predictions than others (\egno, due to occlusion or low quality). Under a framework with per-pixel semantic segmentation and depth regression tasks, input-dependent \emph{heteroscedastic uncertainty} together with \emph{epistemic} uncertainty are considered in new loss functions \cite{kendall2017uncertainties}, making the loss more robust to noisy data. In a multi-task setting, Kendall \etal show that the task uncertainty captures the relative confidence between tasks, reflecting the uncertainty inherent to the regression/classification
task \cite{kendall2018multi}. They propose using \emph{homoscedastic} uncertainty as a basis for weighting losses in a multi-task learning problem. 

In our work, we exploit the \emph{heteroscedastic} uncertainty of the input data (multi-shot images and a corresponding single shot image) to better transfer the knowledge distilled from multi-shot images of an object to each single shot image.

\begin{figure*}
  \centerline{\includegraphics[width=1.0\linewidth]{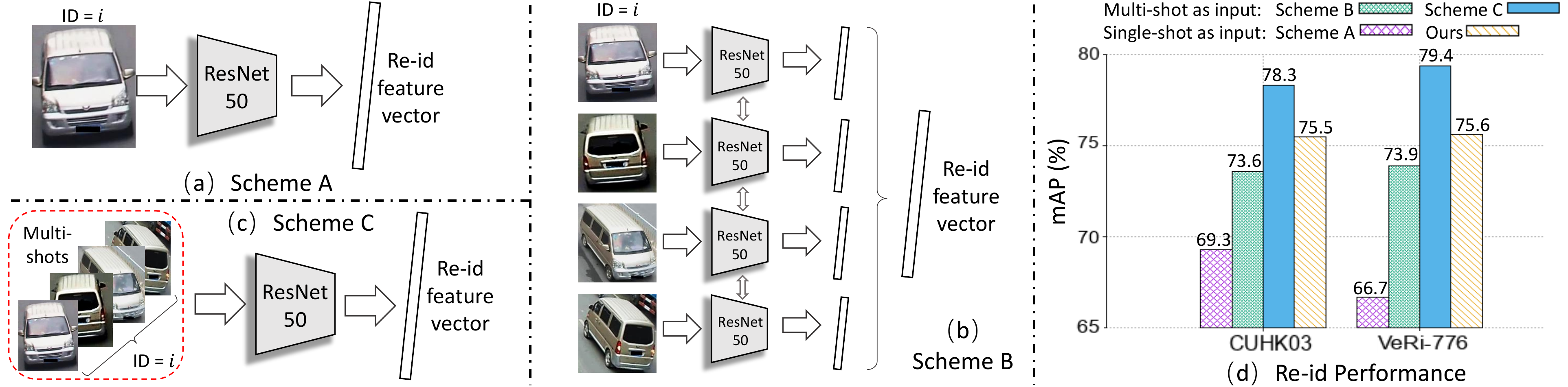}}
  \caption{Investigation on whether using multi-shot images can result in better re-id and how much benefit it potentially brings. (a) \textbf{\emph{Scheme A}} uses single shot image for re-id. (b) \textbf{\emph{Scheme B}} assumes four-shot images are always used and the features of the four images are averaged as the re-id feature. (c) \textbf{\emph{Scheme C}} assumes four-shot images are always used and the re-id feature is jointly extracted from the input of four-shot images (we also use this as the Teacher network in our final scheme). (d) Performance comparisons on  person re-id dataset CUHK03 (labeled setting) and vehicle re-id dataset VeRi-776.}
  \centering
\label{fig:investigation}
\end{figure*}

\section{Uncertainty-aware Multi-shot Teacher-Student (UMTS) Network}\label{sec:UMTS}

We show the proposed Uncertainty-aware Multi-shot Teacher-Student (UMTS) network in Fig. \ref{fig:framework}. It consists of a Teacher network (T-net) that learns comprehensive features from the multi-shot images of the same object, and a Student network (S-net) that takes a single image from this multi-shots as input. We aim to exploit the more comprehensive knowledge from the multi-shots of the same identity to regularize/teach the feature learning of a single image for robust \emph{image-based object re-id}. To effectively transfer knowledge from T-net to S-net, we propose an Uncertainty-aware Knowledge Distillation Loss (UA-KDL) and apply them over intermediate layer features and the final matching features, respectively. The entire network can be trained in an end-to-end manner and only the S-net is needed in inference. We discuss the details in the following subsections.

\subsection{Motivation: Multi-Shots versus Single-Shot}

For an object of the same identity, multiple images captured from different viewpoints/times/places are often available. There are large variations in terms of the visible regions/occlusions, lighting, deformations (\egno, poses of person), and the backgrounds. Multiple images can provide more comprehensive information than a single image. For image-based re-id, each identity usually has multiple images in a dataset even though such grouping information cannot be used in inference, where only a single image is used as the query. There are very few works that explicitly explore the multi-shot information to enhance image-based re-id. 

We look into whether multi-shot images lead to better re-id performance and investigate how much benefit it can bring experimentally. As shown in Fig. \ref{fig:investigation}, we build three schemes (see (a)(b)(c)) based on the ResNet-50 which is widely used in re-id \cite{he2016deep,wang2018learning,wang2018mancs,zhang2019DSA,he2019part}. \emph{Scheme A} is a baseline scheme that uses single image for re-id. \emph{Scheme B} and \emph{Scheme C} both assume four-shots of the same identity \footnote{For each image, based on the groudtruth ids, we randomly select another three images of the same id to have four-shot images.} are used together to obtain the re-id feature vector. \emph{Scheme B} (see (b)) obtains the re-id feature by averaging the feature vectors of the four images while \emph{Scheme C} (see (b)) jointly extracts the re-id feature from the input of four-shot images (input channel number: 3$\times$4). 

We show the performance comparisons on the person re-id dataset CUHK03 (labeled setting) \cite{li2014deepreid} and vehicle re-id dataset VeRi-776 \cite{liu2016deep_VeRI776} in Fig. \ref{fig:investigation}(d). Interestingly, \emph{Scheme B} that simply aggregates the features of four-shots outperforms (\emph{Scheme A}) that uses single image as input by \textbf{4.3\%} and \textbf{7.2\%} in mAP accuracy on CUHK03 and VeRi-776 respectively. \emph{Scheme B} ignores the joint feature extraction and interaction among images of the same id and  \emph{Scheme C} remedies these by simply concatenating four images together in channel as the input. \emph{Scheme C} outperforms \emph{Scheme A} significantly by \textbf{9.0\%} and \textbf{12.7\%} in mAP accuracy on CUHK03 and VeRi-776 respectively. We conclude that there is a huge space for improvement when multi-shot images are available. However, during the inference, in practice, only a single query image is accessible and there is no identity information either for each image in the gallery dataset for image-based re-id. \emph{Scheme B} and \emph{Scheme C} need to take multi-shot images as input and are thus not practical, but somewhat provide performance upper bounds.   

To remedy the practical gap, we propose an Uncertainty-aware Multi-shot Teacher-Student (UMTS) Network to transfer the knowledge of multi-shot images to an individual image (see Fig. \ref{fig:framework}). In inference, as shown in Fig. \ref{fig:investigation}(d), our final scheme UMTS with the S-net alone (Ours), which takes a single image as input, significantly outperforms \emph{Scheme A} by \textbf{6.2\%} and \textbf{8.9\%} in mAP on CUHK03 and VeRi-776 respectively. Note that the model sizes of the three schemes and our final model S-net are almost the same, with \emph{Scheme C} slightly larger (0.11\%), which is fair for comparisons.

\subsection{Teacher network and Student network}\label{subsec:T-net}

Based on the analysis in subsec. 3.1, we take a simple but effective network as in \emph{Scheme C} (see Fig. \ref{fig:investigation}(c)) as the Teacher network (T-net) (see Fig. \ref{fig:framework}). More generally, we define the number of shots as $K$ and the input to the T-net is a tensor of size $H\times W \times 3K$. T-net and S-net have the same network structure beside the difference in the number of the input channels. Each network has four stages with each containing multiple convolutional layers, and one fully connected layer, \ieno, the fifth stage to obtain the re-id feature vector for matching. For each network, we add the widely-used re-identification loss (ReID Loss) (classification loss \cite{sun2017beyond,fu2019horizontal}, and triplet loss with batch hard mining \cite{hermans2017defense}) on the re-id feature vectors. Note that our approach is general and any other networks, \egno, PCB \cite{sun2017beyond}, OSNet \cite{zhou2019omni}, can replace the teacher and student networks. 

For ease of description, we mathematically formulate the construction of training samples for the T-net and S-net. Given the available images of identity $i$, we randomly select $K$ images and obtain the set of $K$-shot images $\mathcal{S}_i = \{\mathcal{X}_{i,1},\cdots, \mathcal{X}_{i,K}\}$, where $\mathcal{X}_{i,k} \in \mathbb{R}^{H \times W \times 3}$. We obtain an input sample for the T-net by concatenating the $K$-shots in channel as $\mathcal{Y}_i \in \mathbb{R}^{H \times W \times 3K}$. $\langle \mathcal{Y}_i$, $\mathcal{X}_{i,k}\rangle$ forms a teacher-student training pair, where $k=1,\cdots, K$. Different from previous teacher-student networks that share the same input data, an input data to our student network (\ieno, $\mathcal{X}_{i,k}$) is only part of the input data to our teacher network (\ieno, $\mathcal{Y}_i$).

\subsection{Knowledge Transfer with UA-KDL}\label{subsec:UA-KDL}

Re-id aims to learn discriminative feature vectors for matching, \egno, in terms of $l_2$ distance. We expect the S-net to learn a representation that is predictive of the representation of the T-net, at both the intermediate layers and the final re-id features. At an intermediate feature level, considering that the inputs to the T-net and S-net $\langle \mathcal{Y}_i$, $\mathcal{X}_{i,k}\rangle$ are different with spatial misalignment in contents, the intermediate feature maps are spatially average pooled to obtain a feature vector before the regularization supervision. We denote $y_i^b = \phi_t^b(\mathcal{Y}_i) \in \mathbb{R}^{c_b}$ as the feature vector at stage $b$ of the T-net with the input $\mathcal{Y}_i$, where $b=1,\cdots,5$. Similarly, we denote $x_{i,k}^b = \phi_s^b(\mathcal{X}_{i,k}) \in \mathbb{R}^{c_b}$ as the feature vector at stage $b$ of the S-net with the input $\mathcal{X}_{i,k}$. We encourage the S-net with a single-shot as input to learn/predict the more comprehensive information from the T-net with an input of $K$-shot images by minimizing the knowledge distillation loss as
\begin{equation}
    \mathcal{L}_{KD(i,k)}^b =  \Vert \theta_t^b(y_i^b) - \theta_s^b(x_{i,k}^b) \Vert ^2, 
    \label{eq:KD-loss}
\end{equation}
where $\theta_t^b(y_i^b)$ and $\theta_s^b(x_{i,k}^b)$ denote projection functions that embed the feature vectors $y_i^b$ and $x_{i,k}^b$ of stage $b$ of the T-net and S-net to the same space/domain. Here $\theta_t^b(y_i^b) = ReLU(BN(W_t^b y_i^b))$ and $\theta_s^b(x_{i,k}^b) = ReLU(BN(W_s^b x_{i,k}^b))$, which is achieved by a fully-connected layer with matrix $W_t^b \in \mathbb{R}^{c_b \times c_b/r_b}$ or $W_s^b \in \mathbb{R}^{c_b \times c_b/r_b}$ followed by a Batch Normalization and ReLU activation function. $r_b$ denotes dimension reduction ratio to reduce the model complexity and aid generalisation. We set $r_b$=16 for $b=1,\cdots,4$, and $r_5$=4 experimentally. Based on the analysis in subsec. 3.1, we can assume the T-net is always better than the S-net in terms of the feature representations. Thus the projected feature $\theta_t^b(y_i^b)$ of the T-net can be considered as the regression target of the S-net. Besides the updating of projection functions $\theta_t^b(\cdot)$ and $\theta_t^b(\cdot)$, the loss are only back-propagated to the S-net to regularize its feature learning as illustrated in Fig. \ref{fig:framework}.

Considering that the $K$ samples of the S-net $\mathcal{S}_i = \{\mathcal{X}_{i,1},\cdots, \mathcal{X}_{i,K}\}$ correspond to the teacher with the same $K$-shot input $\mathcal{Y}_i$, we can optimize the S-net simultaneously from the $K$ samples with the knowledge distillation loss as
\begin{equation}
    \mathcal{L}_{KD(i,:)}^b =  \sum_{k=1}^K\Vert \theta_t^b(y_i^b) - \theta_s^b(x_{i,k}^b) \Vert ^2. 
    \label{eq:KD-loss-K}
\end{equation}
However, in the above formulation, the \emph{heteroscedastic uncertainty} of each sample to approach the features of the T-net is overlooked, where S-net's samples are equally treated.

\textbf{\emph{Heteroscedastic uncertainty}} has been studied from the Bayesian viewpoint and applied to per-pixel depth regression and semantic segmentation tasks, respectively \cite{kendall2017uncertainties,gal2016uncertainty}. It captures noise inherent in the observations, which depends on the input data.

For re-id, different individual images have different visible object regions, occlusions, image quality, and thus have different capability/uncertainty {in acquiring/approaching} the knowledge of the given $K$-shot images of the same identity. Motivated by the uncertainty analysis in Bayesian deep learning and its application in depth regression \cite{kendall2017uncertainties}, we design an Uncertainty-aware Knowledge Distillation Loss (UA-KDL) as
\begin{equation}
    \begin{aligned}
    \mathcal{L}_{UKD(i,:)}^b =  &\sum_{k=1}^K  \frac{1}{{2\sigma_{b}(y_i^b,x_{i,k}^b})^2} \Vert \theta_t^b(y_i^b) - \theta_s^b(x_{i,k}^b) \Vert ^2 \\
    &+\log\sigma_{b}(y_i^b,x_{i,k}^b) ,
    \end{aligned}
    \label{eq:UKD-loss-K}
\end{equation}
where $\sigma_{b}(y_i^b,x_{i,k}^b)$ denotes the observation noise parameter for capturing \emph{heteroscedastic uncertainty} in regression and is data-dependent. Based on the uncertainty analysis in \cite{kendall2017uncertainties}, minimizing this loss actually is equivalent to maximizing the log likelihood of the regression for the purpose of approaching the feature of T-net by the S-net $p(\theta_t^b(y_i^b)| \theta_s^b(\phi_s^b(\mathcal{X}_{i,k}))$. The introduction of uncertainty factors allows the S-net to adaptively allocate learning efforts on different samples for effectively training the network. For example, for a noisy image with the object being seriously occluded, the uncertainty to approach the feature of multi-shot images is high (\ieno, large $\sigma_b$) and it is wise to give small weight to the loss to have a smaller effect. The second item can prevent predicting infinite uncertainty (and therefore zero loss for the first item) for all images.

In our framework, the \emph{heteroscedastic uncertainty} for regression depends on \emph{both the feature of the} $K$\emph{-shot images (which is the target)} and \emph{the feature of the single image (which intends to approach the target)}. Then we model the log of uncertainty \ieno, $\upsilon_b(y_{i}^{b}, x_{i,k}^{b}) := {\rm log} ({\sigma_b}(y_{i}^{b}, x_{i,k}^{b})^2)$,
\begin{equation}
    \begin{aligned}
        {\upsilon_b}(y_{i}^{b}, x_{i,k}^{b}) = & ReLU({\rm w_b}[\theta_t^b(y_i^b), \theta_s^b(x_{i,k}^b)]),
    \end{aligned}
\end{equation}
where $[\cdot,\cdot]$ denotes the concatenation, $\rm w_b$ is achieved by a fully connected layer to map $[\theta_t^b(y_i^b), \theta_s^b(x_{i,k}^b)]$ to a scalar followed by ReLU. Predicting the log of uncertainty is more numerically stable than predicting ${\sigma_b}$, since this avoids a potential division by zero in (\ref{eq:UKD-loss-K})\cite{kendall2017uncertainties}.

\subsection{Training and Inference}\label{subsec:UA-KDL}

As in Fig. \ref{fig:framework}, for the $K$-shot images of the same identity $i$, the overall optimization loss consists of the widely used re-identification loss $\mathcal{L}_{ReID}$, and the proposed UA-KDLs:
\begin{equation}
    \mathcal{L}_{(i,:)} = \mathcal{L}_{{ReID}_{(i,:)}} +  \sum_{b=1}^5 \lambda_b \mathcal{L}_{UKD(i,:)}^b.
    \label{eq:total-loss}
\end{equation}
Note that we add the UA-KDL at all 5 stages (the first 4 stages and the final re-id feature vector of the last stage) to enable the knowledge transfer on intermediate features and the final re-id features. $\lambda_b$ is a weight to control the relative importance for the regularization at stage $b$. In considering the re-id feature of stage $5$ is more relevant to the task, we experimentally set $\lambda_5=0.5$, and $\lambda_b=0.1$ for the first 4 stages. We find that training the T-net first to convergence and then fixing the T-net followed by the joint training of S-net and UA-KDL related parameters can produce better performance (about 1.4\% gain in mAP on CUHK03(L)) than the end-to-end joint training. {This can all along leverage the stable superior performance of the T-net.}

In inference, we use only the S-net without any increase in computational or model complexity. The feature vector $x_{i,k}^5$ from stage 5 is the final re-id feature for matching.

\section{Experiments}

\subsection{Datasets and Evaluation Metrics}

We conduct object re-id experiments on the most commonly-used person re-id dataset, CUHK03 \cite{li2014deepreid} (including the labeled/detected bounding box settings), and three vehicle re-id datasets of VeRi-776 \cite{liu2016deep_VeRI776}, VehicleID \cite{liu2016deep_VehicleID} and the recent large-scale VERI-Wild \cite{lou2019veri}.

We follow common practices and use the cumulative matching characteristics (CMC) at Rank-1, and mean average precision (mAP) to evaluate the performance.

\subsection{Implementation Details}
We use ResNet-50 \cite{he2016deep} to build the T-net, S-net, and baseline respectively. We set $K$ as 4 and add UA-KDLs at all the 5 stages by default. The batch size is set as 64. Following \cite{hermans2017defense}, a batch is formed by first randomly sampling $P$ identities. For each identity, we then sample $K$ images. Then the batch size is $P\times K$ for the S-net and $P$ for the T-net. For simplicity, we refer to batch size with respect to the S-net hereafter. The input image resolution is set to 256$\times$256 for vehicle re-id and 256$\times$128 for person re-id, respectively. 

We use the commonly used data augmentation strategies of random cropping \cite{zhang2019DSA}, horizontal flipping, label smoothing regularization \cite{szegedy2016rethinking}, and random erasing \cite{zhong2017random} in both the baseline schemes and our schemes. We use Adam optimizer \cite{kingma2014adam} for model optimization. All our models are implemented on PyTorch and trained on a single NVIDIA-P40 GPU.

\subsection{Ablation Study}
We perform comprehensive ablation studies to demonstrate the effectiveness of the designs in our UMTS framework, on both the person re-id dataset CUHK03 (labeled bounding box setting) and the vehicle re-id dataset VeRi-776.

\noindent\textbf{Effectiveness of Our Framework.}
Table \ref{tab:Baseline-UMTS} shows the comparisons of our schemes with the baseline. \textbf{\emph{Baseline}} denotes the baseline scheme without taking into account multi-shot images. \textbf{\emph{MTS}} denotes our \textbf{M}ulti-shot \textbf{T}eacher-\textbf{S}tudent Network with knowledge distillation \emph{without} considering the \emph{heteroscedastic uncertainty} (see formulation (\ref{eq:KD-loss-K})). \textbf{\emph{UMTS}} denotes our final \textbf{U}ncertainty-aware \textbf{M}ulti-shot \textbf{T}eacher-\textbf{S}tudent Network (see formulation (\ref{eq:UKD-loss-K})). \textbf{\emph{UMTS(all)}} denotes that the UA-KDL is applied at all five stages ($b$1 to $b$5). Similarly, \textbf{\emph{UMTS(b5)}} denotes that the UA-KDL is only added at stage 5 while there is no knowledge distillation loss on the other 4 stages. We make the following observations/conclusions.
\begin{enumerate}[label=\arabic*),leftmargin=*,noitemsep,nolistsep]

\item Thanks to the exploration of the knowledge from multi-shot images, and the \emph{heteroscedastic uncertainty}, our final scheme \emph{UMTS(all)} significantly outperforms \emph{Baseline} by \textbf{6.2\%} and \textbf{8.9\%} in mAP accuracy on CUHK03(L) and VeRi-776, respectively. 

\item By learning the knowledge from multi-shots, our  \emph{MTS(all)} outperforms \emph{Baseline} by \textbf{4.6\%} and \textbf{6.1\%} in mAP on CUHK03(L) and VeRi-776, respectively.

\item \emph{UMTS(all)}, which introduces the \emph{heteroscedastic uncertainty}, further improves the mAP accuracy by \textbf{1.6\%} and \textbf{2.8\%} on CUHK03(L) and VeRi-776, respectively.

\end{enumerate}

\begin{table}[t]
  \centering
  \footnotesize
  \tabcolsep=9.5pt
  \caption{Performance (\%) of our schemes and \emph{Baseline}. \emph{MTS} denotes our Multi-shot Teacher-Student Network. \emph{UMTS} denotes Uncertainty-aware MTS. ($b$m+$b$n) denotes the knowledge distillation losses are applied over stage m and n.}
    \begin{tabular}{ccccc}
    \toprule
    \multirow{2}[4]{*}{Model} & \multicolumn{2}{c}{CUHK03(L)} & \multicolumn{2}{c}{VeRi-776} \\
\cmidrule{2-5}          & Rank-1 & mAP   & Rank-1 & mAP \\
    \midrule
    Baseline  & 73.5  & 69.3  & 91.8  & 66.7 \\
    \midrule
    MTS($b$5) & 74.3  & 71.1  & 92.9      & 69.3 \\
    UMTS($b$5) & 76.6  & 72.8  & 93.7  & 70.9 \\
    \midrule
    UMTS($b$1+$b$5) & 77.8  & 73.4  & 93.9  & 71.8 \\
    UMTS($b$2+$b$5) & 78.6  & 74.3  & 94.4  & 73.5 \\
    UMTS($b$3+$b$5) & 79.3  & 74.8  & 94.8  & 74.1 \\
    UMTS($b$4+$b$5) & 78.8  & 74.5  & 94.4  & 74.0 \\
    \midrule
    MTS(all) & 77.7  & 73.9  & 94.0 & 72.8  \\
    \textbf{UMTS(all)} & \textbf{80.7}  & \textbf{75.5}  & \textbf{95.8}  & \textbf{75.9} \\
    \bottomrule
    \end{tabular}%
  \label{tab:Baseline-UMTS}%
\end{table}%

\subsection{Design Choices of UMTS}

\noindent\textbf{Which Stage to Add UA-KDL?}
To transfer the knowledge from the T-net to the S-net, we add the UA-KDL over the final stage re-id features (which are the most task-relevent) as the scheme \emph{UMTS(b5)}. \emph{UMTS(b5)} outperforms \emph{Baseline} by 3.5\% and 4.2\% in mAP on CUHK03 and VeRi-776, respectively. We compare the cases of adding an UA-KDL to a different stage (Conv Block), and adding UA-KDLs to all stages (\ieno, see Fig. \ref{fig:framework}). Table \ref{tab:Baseline-UMTS} shows the results. We observe that on each stage, the adding of UA-KDL leads to obvious improvement and the gains are larger on stages 3, 4 and 2. When UA-KDL are added to all 5 stages, our scheme \emph{UMTS(all)} achieves the best performance.

\begin{figure}
  \centerline{\includegraphics[width=1.0\linewidth]{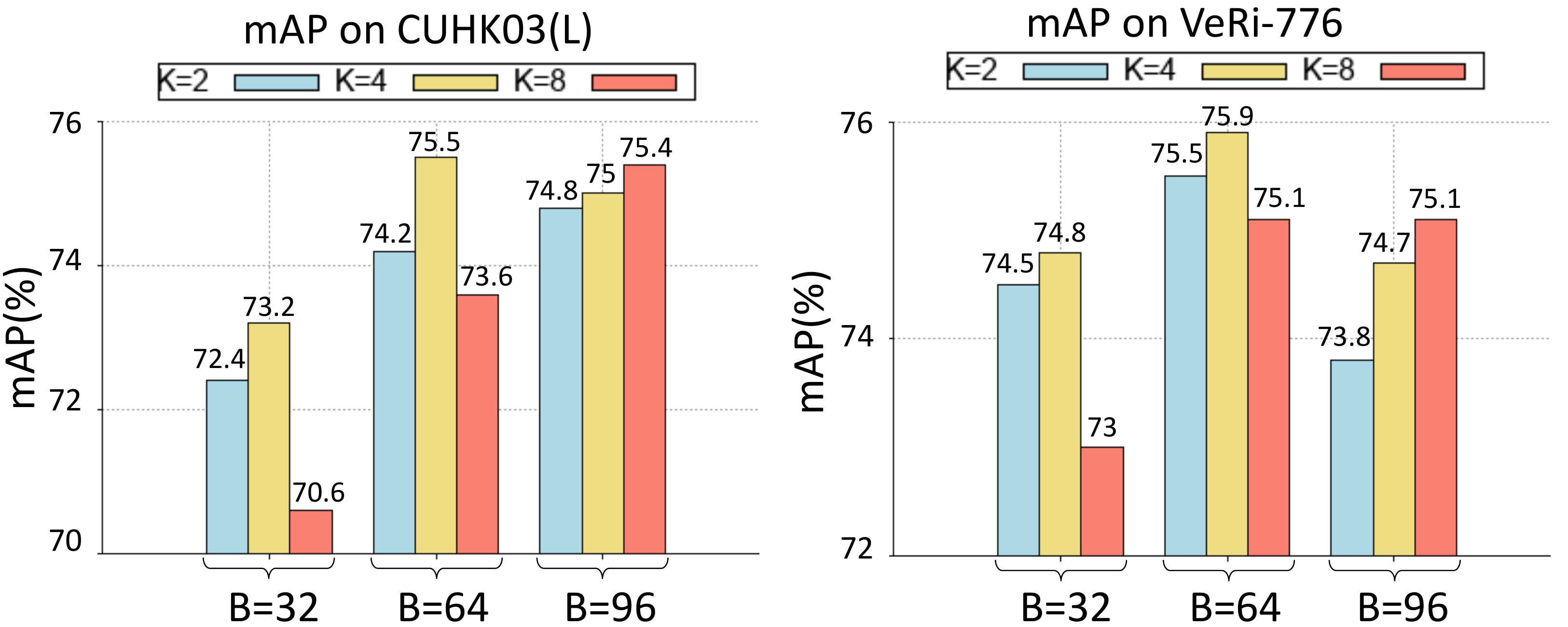}}
  \caption{Study on the number of shots $K$ and batch size $B$.}
  \centering
\label{fig:NK}
\end{figure}

\noindent\textbf{Influence of the Number of Shots $K$ and Batch Size $B$.}
We study the influence of the number of shots $K$ ($K$=2,~4,~8) on re-id performance under the settings of different batch sizes $B$ ($B$=32,~64,~96 which is commonly used in re-id) on CUHK03 and VeRi-776 datasets and show the results in Fig. \ref{fig:NK}. We have the following observations.

\begin{enumerate}[label=\arabic*),leftmargin=*,noitemsep,nolistsep]
\item For batch size $B$=64, the setting with $K$=4 shots provides the best performance. That may be because a smaller number of shots (\egno, $K$=2) for the T-net cannot provide enough comprehensive information. When the shot number is too large, \egno, $K$=8, the number of samples (equivalent batch size) for the T-net is small, \egno, $B/K$=8, which is not enough to have precise statistical estimation for the Batch Normalization layers. 
\item When the batch size is small, {\ieno}, $B$=32, large shot number $K$=8 results in inferior performance because of too small number ($B/K$=4) of samples in a batch for the T-net. When the batch size is increased to $B$=96, the performance for $K$=8 shots further increases on CUHK03 and saturates on VeRi-776. For $K$=4, the increase of batch size does not bring benefit since using large batch tends to converge to sharp minimums and leads to poorer generalization \cite{keskar2017large}. Besides, too large batch size requires significant GPU memory resources. 
\end{enumerate}
\emph{We set $B$=64 and $K=4$ to trade off the performance and GPU memory requirement.}

\begin{figure}
  \centerline{\includegraphics[width=1.0\linewidth]{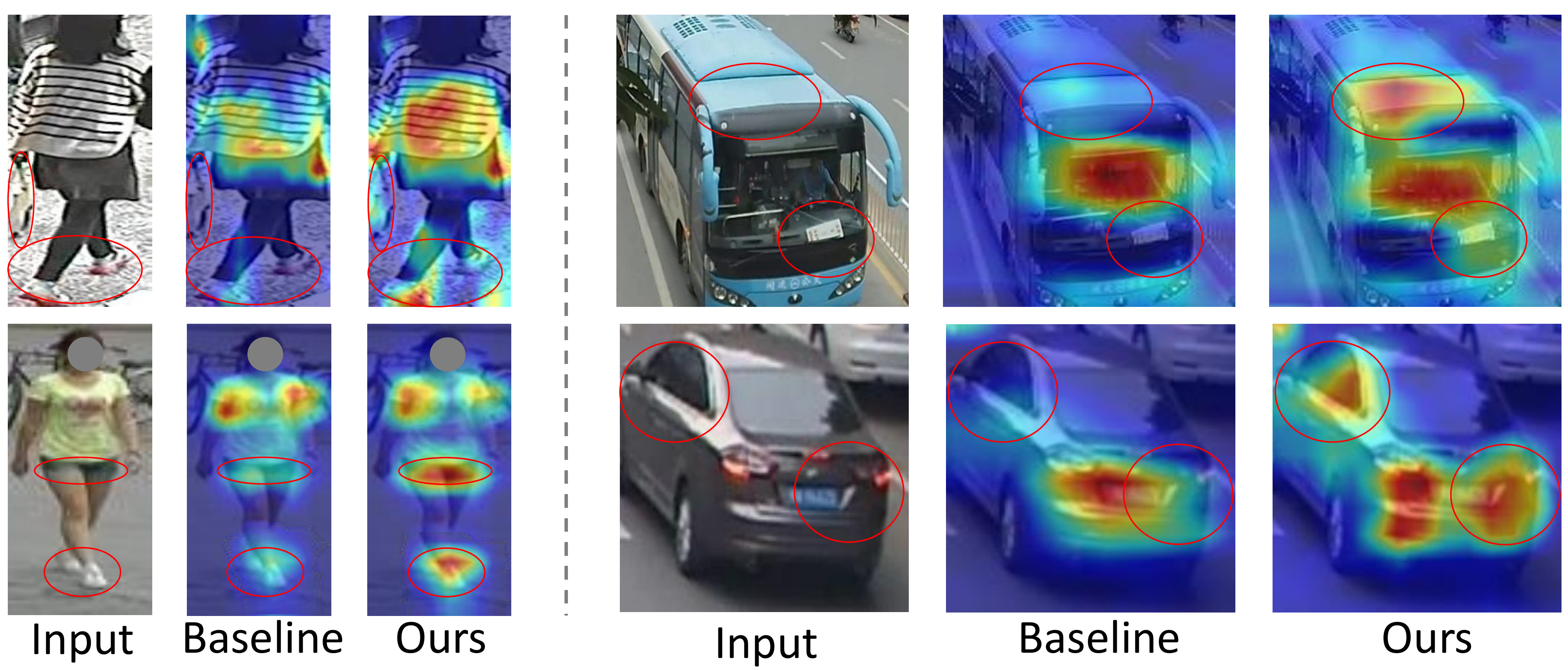}}
  \caption{Gradient responses from \emph{Baseline} and our S-net (\emph{Ours}) using the tool Grad-CAM. Best viewed in color.}
  \centering
\label{fig:att}
\end{figure}

\begin{figure}
  \centerline{\includegraphics[width=1.0\linewidth]{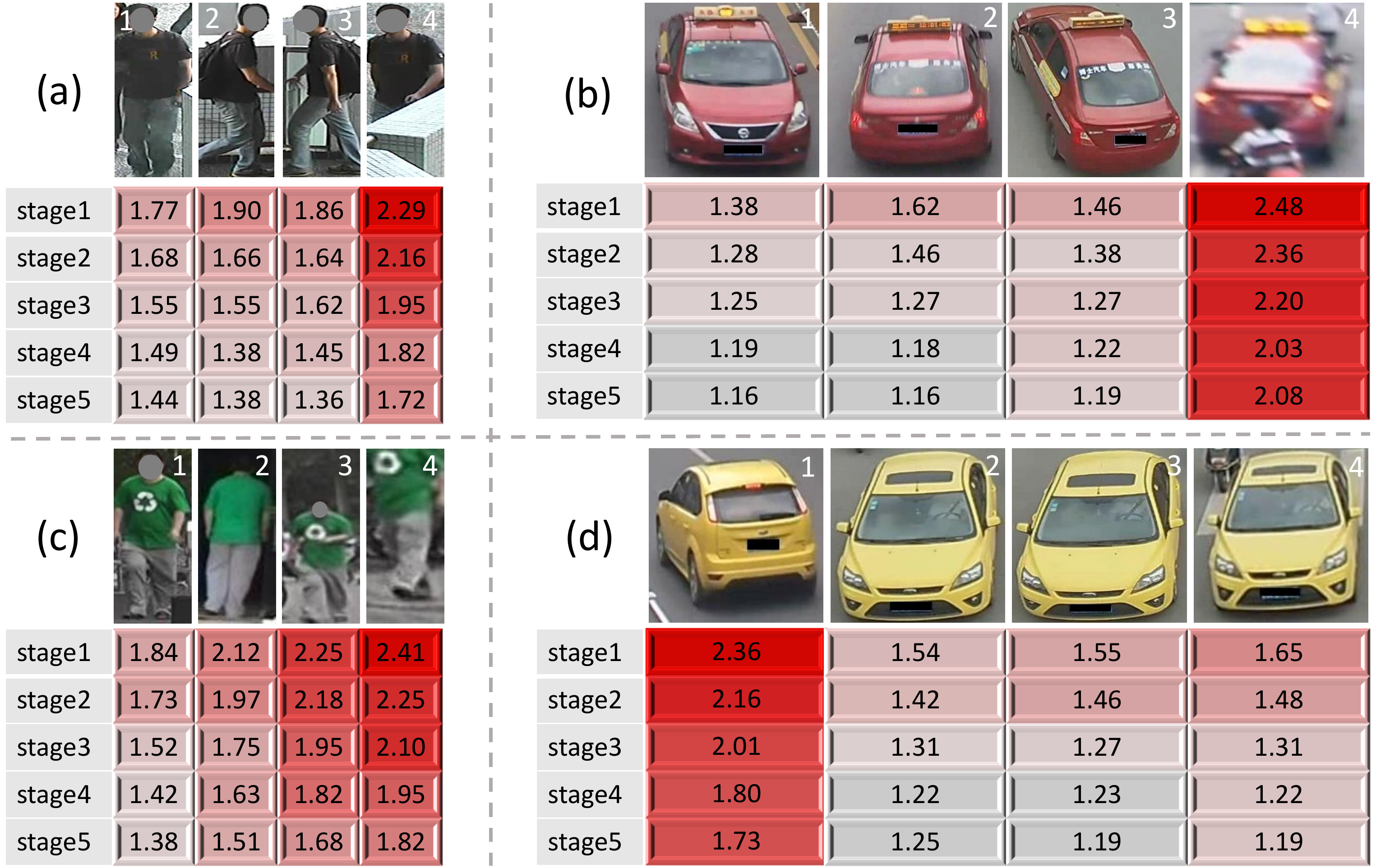}}
  \caption{Predicted uncertainty $\sigma_b^2$ for the 4-shot images on the 5 stages respectively. Red represents large weight and white means small weight. Best viewed in color.}
  \centering
\label{fig:U}
\end{figure}

\begin{table*}
    \centering
    \tiny
    \caption{Performance (\%) comparsions of the proposed UMTS and state-of-the-art methods on the vehicle re-id datasets.}
    \begin{tabular}{lcccccccccccccc}
    \toprule
    \multicolumn{1}{c}{\multirow{3}[6]{*}{Methods}} & \multicolumn{2}{c}{\multirow{2}[4]{*}{VeRi-776}} & \multicolumn{6}{c}{VehicleID}                 & \multicolumn{6}{c}{VERI-Wild} \\
\cmidrule{4-15}          & \multicolumn{2}{c}{} & \multicolumn{2}{c}{Small=800} & \multicolumn{2}{c}{Medium=1600} & \multicolumn{2}{c}{Large=2400} & \multicolumn{2}{c}{Small} & \multicolumn{2}{c}{Medium} & \multicolumn{2}{c}{Large} \\
\cmidrule{2-15}          & Rank-1 & mAP   & Rank-1 & mAP   & Rank-1 & mAP   & Rank-1 & mAP   & Rank-1 & mAP   & Rank-1 & mAP   & Rank-1 & mAP \\
    \midrule
    GSTE (TMM)~\cite{bai2018group} & --      & --      & --      & --      & --      & --      & --      & --      & 60.4 & 31.4 & 52.1 & 26.1 & 45.3 & 19.5 \\
    VAMI (CVPR)~\cite{zhou2018aware} & 77.0 & 50.1 & 63.1  & --      & 52.8 & --      & 47.3 & --      &  --     & --      &  --     & --      & --      & -- \\
    FDA-Net (CVPR)~\cite{lou2019veri} & 84.2 & 55.4 & --      & --      & 59.8 & 65.3 & 55.5 & 61.8 & 64.0 & 35.1 & 57.8 & 29.8  & 49.4 & 22.7 \\
    Part-regularized (CVPR)~\cite{he2019part} & \underline{94.3}  & \underline{74.3}  & \underline{78.4}  & --      & \underline{75}    &  --     & \underline{74.2}  & --      & --      & --      & --      & --      & --      & -- \\
    MoV1+BS (IJCNN)~\cite{kumar2019vehicle} & 90.2  & 67.6  & 78.2  & \underline{86.1}  & --      & \underline{81.7}  & --      & \underline{78.2}  & \underline{82.9}  & \underline{68.7} & \underline{77.6} & \underline{61.1} & \underline{69.5} & \underline{49.7} \\
    \midrule
    Baseline & 91.8  & 66.7  & 74.4  & 80.4  & 72.4  & 77.1  & 69.8  & 75.2  & 77.6  & 65.2  & 72.7  & 60.3  & 63.8  & 45.0 \\
    UMTS  & \textbf{95.8} & \textbf{75.9} & \textbf{80.9} & \textbf{87.0} & \textbf{78.8} & \textbf{84.2} & \textbf{76.1} & \textbf{82.8} & \textbf{84.5} & \textbf{72.7} & \textbf{79.3} & \textbf{66.1} & \textbf{72.8} & \textbf{54.2} \\
    \bottomrule
\end{tabular}
\label{tab:STO-vehicle}
\end{table*}

\subsection{Visualization}

\textbf{Visualization of Feature Maps.} To understand how the multi-shot images benefit the feature learning of the single image, we use Grad-CAM \cite{selvaraju2017grad} to visualize the gradient responses of \emph{Baseline} and S-net of our \emph{UMTS} in Fig. \ref{fig:att}. We observe that \emph{Baseline} tends to pay more attention to some local regions and ignore some potential discriminative regions on an object. In contrast, by exploiting knowledge from multi-shot images which have a more comprehensive perspective, our S-net can pay attention to more regions to capture more discriminative information, such as  `bag', `shoes' (first row), and `shorts' (second row) on the persons, and `inspection sticker' on the bus.

\noindent\textbf{Visualization of Learned Uncertainty $\sigma_b^2$.}
We visualize the predicted uncertainty factor $\sigma_b^2$ for the $K$-shot images ($K$=4) on the five stages respectively in Fig. \ref{fig:U}. In (a) and (b), the uncertainties of image-4 ($I_4$) are both relatively large due to occlusion and poor image quality (blur). In (c), the uncertainty values of $I_3$ and $I_4$ are larger than that of $I_1$ and $I_2$, which may be because of the small scale and incompleteness of the person. Fig. \ref{fig:U} (d) shows that $I_1$ belongs to the minority and thus has the highest uncertainty while the other similar images have similar low uncertainty.

\subsection{Comparison with State-of-the-Arts}
We compare the proposed UMTS with other state-of-the-art approaches and show the results in Table \ref{tab:STO-vehicle} and Table \ref{tab:STO-person} for vehicle re-id and person re-id, respectively. With the same network structure in inference, our UMTS significantly outperforms \emph{Baseline(ResNet-50)} on all the datasets, by \textbf{9.2\%} and \textbf{7.2\%} in mAP on the large-scale VERI-Wild for vehicle re-id, and CUHK03 (labeled) for person re-id, respectively. Our approach achieves the best performance on \emph{all} the vehicle re-id datasets and most of the person re-id datasets. On the large-scale vehicle dataset VERI-Wild, our approach outperforms the second best approach by 4.0\%, 5.0\%, 4.5\% for small, medium, and large sub-test sets, respectively.

\begin{table}
    \centering
    \tiny
    \tabcolsep=6.8pt
   \caption{Performance (\%) comparsions of UMTS and state-of-the-art methods on the person re-id dataset CUHK03.}
    \begin{tabular}{lcccc}
    \toprule
    \multicolumn{1}{c}{\multirow{3}[6]{*}{Method}} & \multicolumn{4}{c}{CUHK03} \\
\cmidrule{2-5}          & \multicolumn{2}{c}{Labeled} & \multicolumn{2}{c}{Detected} \\
\cmidrule{2-5}          & Rank-1 & mAP   & Rank-1 & mAP \\
    \midrule
    HA-CNN (CVPR)~\cite{li2018harmonious} & 44.4  & 41.0  & 41.7  & 38.6 \\
    PCB+RPP (ECCV)~\cite{sun2017beyond} & 63.7  & 57.5  & --      & --     \\
    Mancs (ECCV)~\cite{wang2018mancs} & 69.0  & 63.9  & 65.5  & 60.5  \\
    MGN (ACMMM)~\cite{wang2018learning} & 68.0  & 67.4  & 66.8  & 66.0  \\
    HPM (AAAI)~\cite{fu2019horizontal} & 63.9  & 57.5  & --      & --     \\
    CAMA (CVPR)~\cite{yang2019towards} & 70.1  & 66.5  & 66.6  & 64.2  \\    
    CASN (CVPR)~\cite{zheng2019re} & 73.7  & 68.0  & 71.5  & 64.4 \\    
    DSA-reID (CVPR)~\cite{zhang2019DSA} & \underline{78.9}  & \underline{75.2}  & \underline{78.2} & 73.1  \\  
    \midrule
    Baseline (ResNet-50) & 73.5  & 69.3  & 70.0  & 66.0  \\
    UMTS (ResNet-50)  & \textbf{80.7} & \textbf{75.5} & 77.2  & \underline{73.4} \\
    \midrule
    Baseline (OSNet) (ICCV)~\cite{zhou2019omni} & --  & --  & 72.3  & 67.8 \\
    UMTS (OSNet) & --	& --	& \textbf{78.6}	& \textbf{74.1} \\
    \bottomrule
\end{tabular}
\label{tab:STO-person}
\end{table}

Besides, our proposed UMTS is a general framework, and can be easily applied to other powerful backbone networks to achieve superior performance. Table \ref{tab:STO-person} also shows the comparison when using OSNet \cite{zhou2019omni} as the backbone for person re-id. In comparison with this superior baseline (which outperforms ResNet-50 backbone by 1.8\% in mAP), our UMTS achieves 6.3\% improvement in mAP, and achieves the best performance. 

\section{Conclusion}

In this paper, we propose a simple yet powerful Uncertainty-aware Multi-shot Teacher-Student (UMTS) framework to exploit the comprehensive information of multi-shot images of the same identity for effective single image based re-id. In particular, to efficiently transfer knowledge from the T-net to S-net, we take into account the \emph{heteroscedastic uncertainty} related to the single image input to the S-net and the $K$-shot images input to the T-net and design an Uncertainty-aware Knowledge Distillation Loss (UA-KDL) which is applied at different semantics levels/stages. Extensive experiments on person re-id and vehicle re-id both demonstrate the effectiveness of the designs. Our UMTS achieves the best performance on all the three vehicle re-id datasets and the person re-id dataset. In inference, we only use the S-net without any increase in computational cost and model complexity.


\section{Acknowledgments}
This work was supported in part by NSFC under Grant 61571413, 61632001.
We would like to thank Yizhou Zhou for the valuable discussion.

{
\small
\bibliographystyle{aaai}
\bibliography{reference}

\begin{thebibliography}{}

\bibitem[\protect\citeauthoryear{Bai \bgroup et al\mbox.\egroup
  }{2018}]{bai2018group}
Bai, Y.; Lou, Y.; Gao, F.; et~al.
\newblock 2018.
\newblock Group-sensitive triplet embedding for vehicle reidentification.
\newblock {\em IEEE TMM} 20(9):2385--2399.

\bibitem[\protect\citeauthoryear{Chen \bgroup et al\mbox.\egroup
  }{2017}]{chen2017learning}
Chen, G.; Choi, W.; Yu, X.; et~al.
\newblock 2017.
\newblock Learning efficient object detection models with knowledge
  distillation.
\newblock In {\em NeurIPS}.

\bibitem[\protect\citeauthoryear{Fu \bgroup et al\mbox.\egroup
  }{2019}]{fu2019horizontal}
Fu, Y.; Wei, Y.; Zhou, Y.; et~al.
\newblock 2019.
\newblock Horizontal pyramid matching for person re-identification.
\newblock In {\em AAAI}, volume~33,  8295--8302.

\bibitem[\protect\citeauthoryear{Gal}{2016}]{gal2016uncertainty}
Gal, Y.
\newblock 2016.
\newblock {\em Uncertainty in deep learning}.
\newblock Ph.D. Dissertation, PhD thesis, University of Cambridge.

\bibitem[\protect\citeauthoryear{Ge \bgroup et al\mbox.\egroup
  }{2018}]{ge2018fd}
Ge, Y.; Li, Z.; Zhao, H.; et~al.
\newblock 2018.
\newblock Fd-gan: Pose-guided feature distilling gan for robust person
  re-identification.
\newblock In {\em NeurIPS}.

\bibitem[\protect\citeauthoryear{He \bgroup et al\mbox.\egroup
  }{2016}]{he2016deep}
He, K.; Zhang, X.; Ren, S.; et~al.
\newblock 2016.
\newblock Deep residual learning for image recognition.
\newblock In {\em CVPR},  770--778.

\bibitem[\protect\citeauthoryear{He \bgroup et al\mbox.\egroup
  }{2019}]{he2019part}
He, B.; Li, J.; Zhao, Y.; and Tian, Y.
\newblock 2019.
\newblock Part-regularized near-duplicate vehicle re-identification.
\newblock In {\em CVPR},  3997--4005.

\bibitem[\protect\citeauthoryear{Hermans, Beyer, and
  Leibe}{2017}]{hermans2017defense}
Hermans, A.; Beyer, L.; and Leibe, B.
\newblock 2017.
\newblock In defense of the triplet loss for person re-identification.
\newblock {\em arXiv preprint arXiv:1703.07737}.

\bibitem[\protect\citeauthoryear{Hinton, Vinyals, and
  Dean}{2015}]{hinton2015distilling}
Hinton, G.; Vinyals, O.; and Dean, J.
\newblock 2015.
\newblock Distilling the knowledge in a neural network.
\newblock {\em arXiv preprint arXiv:1503.02531}.

\bibitem[\protect\citeauthoryear{Huang \bgroup et al\mbox.\egroup
  }{2018}]{huang2018multi}
Huang, Y.; Xu, J.; Wu, Q.; et~al.
\newblock 2018.
\newblock Multi-pseudo regularized label for generated data in person
  re-identification.
\newblock {\em TIP} 28(3).

\bibitem[\protect\citeauthoryear{Kalayeh \bgroup et al\mbox.\egroup
  }{2018}]{kalayeh2018human}
Kalayeh, M.~M.; Basaran, E.; G{\"o}kmen, M.; et~al.
\newblock 2018.
\newblock Human semantic parsing for person re-identification.
\newblock In {\em CVPR}.

\bibitem[\protect\citeauthoryear{Kendall and
  Gal}{2017}]{kendall2017uncertainties}
Kendall, A., and Gal, Y.
\newblock 2017.
\newblock What uncertainties do we need in bayesian deep learning for computer
  vision?
\newblock In {\em NeurIPS}.

\bibitem[\protect\citeauthoryear{Kendall, Gal, and
  Cipolla}{2018}]{kendall2018multi}
Kendall, A.; Gal, Y.; and Cipolla, R.
\newblock 2018.
\newblock Multi-task learning using uncertainty to weigh losses for scene
  geometry and semantics.
\newblock In {\em CVPR},  7482--7491.

\bibitem[\protect\citeauthoryear{Keskar \bgroup et al\mbox.\egroup
  }{2017}]{keskar2017large}
Keskar, N.~S.; Mudigere, D.; Nocedal, J.; et~al.
\newblock 2017.
\newblock On large-batch training for deep learning: Generalization gap and
  sharp minima.
\newblock In {\em ICLR}.

\bibitem[\protect\citeauthoryear{Kingma and Ba}{2014}]{kingma2014adam}
Kingma, D.~P., and Ba, J.
\newblock 2014.
\newblock Adam: A method for stochastic optimization.
\newblock {\em International Conference on Learning Representations}.

\bibitem[\protect\citeauthoryear{Kumar \bgroup et al\mbox.\egroup
  }{2019}]{kumar2019vehicle}
Kumar, R.; Weill, E.; Aghdasi, F.; et~al.
\newblock 2019.
\newblock Vehicle re-identification: an efficient baseline using triplet
  embedding.
\newblock {\em IJCNN}.

\bibitem[\protect\citeauthoryear{Li \bgroup et al\mbox.\egroup
  }{2014}]{li2014deepreid}
Li, W.; Zhao, R.; Tian, L.; et~al.
\newblock 2014.
\newblock Deepreid: Deep filter pairing neural network for person
  re-identification.
\newblock In {\em CVPR},  152--159.

\bibitem[\protect\citeauthoryear{Li \bgroup et al\mbox.\egroup
  }{2017}]{li2017learning}
Li, D.; Chen, X.; Zhang, Z.; et~al.
\newblock 2017.
\newblock Learning deep context-aware features over body and latent parts for
  person re-identification.
\newblock In {\em CVPR}.

\bibitem[\protect\citeauthoryear{Li, Zhu, and Gong}{2018}]{li2018harmonious}
Li, W.; Zhu, X.; and Gong, S.
\newblock 2018.
\newblock Harmonious attention network for person re-identification.
\newblock In {\em CVPR}.

\bibitem[\protect\citeauthoryear{Liu \bgroup et al\mbox.\egroup
  }{2016}]{liu2016deep_VeRI776}
Liu, X.; Liu, W.; Yang, Y.; et~al.
\newblock 2016.
\newblock A deep learning-based approach to progressive vehicle
  re-identification for urban surveillance.
\newblock In {\em ECCV},  869--884.

\bibitem[\protect\citeauthoryear{Liu \bgroup et al\mbox.\egroup
  }{2017}]{liu2017end}
Liu, H.; Feng, J.; Qi, M.; et~al.
\newblock 2017.
\newblock End-to-end comparative attention networks for person
  re-identification.
\newblock {\em TIP}  3492--3506.

\bibitem[\protect\citeauthoryear{Liu \bgroup et al\mbox.\egroup
  }{2018a}]{liu2018pose}
Liu, J.; Ni, B.; Zhuang, Y.; et~al.
\newblock 2018a.
\newblock Pose transferrable person re-identification.
\newblock In {\em CVPR}.

\bibitem[\protect\citeauthoryear{Liu \bgroup et al\mbox.\egroup
  }{2018b}]{RAM2018}
Liu, X.; Zhang, S.; Huang, Q.; et~al.
\newblock 2018b.
\newblock Ram: A region-aware deep model for vehicle re-identification.
\newblock In {\em ICME}.

\bibitem[\protect\citeauthoryear{Liu, Tian, and
  others}{2016}]{liu2016deep_VehicleID}
Liu, H.; Tian, Y.; et~al.
\newblock 2016.
\newblock Deep relative distance learning: Tell the difference between similar
  vehicles.
\newblock In {\em CVPR},  2167--2175.

\bibitem[\protect\citeauthoryear{Lou \bgroup et al\mbox.\egroup
  }{2019}]{lou2019veri}
Lou, Y.; Bai, Y.; Liu, J.; et~al.
\newblock 2019.
\newblock Veri-wild: A large dataset and a new method for vehicle
  re-identification in the wild.
\newblock In {\em CVPR}.

\bibitem[\protect\citeauthoryear{Qian \bgroup et al\mbox.\egroup
  }{2018}]{qian2018pose}
Qian, X.; Fu, Y.; Wang, W.; et~al.
\newblock 2018.
\newblock Pose-normalized image generation for person re-identification.
\newblock In {\em ECCV}.

\bibitem[\protect\citeauthoryear{Romero \bgroup et al\mbox.\egroup
  }{2015}]{romero2015fitnets}
Romero, A.; Ballas, N.; Kahou, S.~E.; et~al.
\newblock 2015.
\newblock Fitnets: Hints for thin deep nets.
\newblock In {\em ICLR}.

\bibitem[\protect\citeauthoryear{Selvaraju \bgroup et al\mbox.\egroup
  }{2017}]{selvaraju2017grad}
Selvaraju, R.~R.; Cogswell, M.; Das, A.; et~al.
\newblock 2017.
\newblock Grad-cam: Visual explanations from deep networks via gradient-based
  localization.
\newblock In {\em ICCV},  618--626.

\bibitem[\protect\citeauthoryear{Su \bgroup et al\mbox.\egroup
  }{2017}]{su2017pose}
Su, C.; Li, J.; Zhang, S.; et~al.
\newblock 2017.
\newblock Pose-driven deep convolutional model for person re-identification.
\newblock In {\em ICCV}.

\bibitem[\protect\citeauthoryear{Subramaniam, Chatterjee, and
  Mittal}{2016}]{subramaniam2016deep}
Subramaniam, A.; Chatterjee, M.; and Mittal, A.
\newblock 2016.
\newblock Deep neural networks with inexact matching for person
  re-identification.
\newblock In {\em NeurIPS},  2667--2675.

\bibitem[\protect\citeauthoryear{Suh \bgroup et al\mbox.\egroup
  }{2018}]{suh2018part}
Suh, Y.; Wang, J.; Tang, S.; et~al.
\newblock 2018.
\newblock Part-aligned bilinear representations for person re-identification.
\newblock In {\em ECCV}.

\bibitem[\protect\citeauthoryear{Sun \bgroup et al\mbox.\egroup
  }{2018}]{sun2017beyond}
Sun, Y.; Zheng, L.; Yang, Y.; et~al.
\newblock 2018.
\newblock Beyond part models: Person retrieval with refined part pooling.
\newblock In {\em ECCV}.

\bibitem[\protect\citeauthoryear{Szegedy \bgroup et al\mbox.\egroup
  }{2016}]{szegedy2016rethinking}
Szegedy, C.; Vanhoucke, V.; Ioffe, S.; Shlens, J.; and Wojna, Z.
\newblock 2016.
\newblock Rethinking the inception architecture for computer vision.
\newblock In {\em CVPR}.

\bibitem[\protect\citeauthoryear{Wang \bgroup et al\mbox.\egroup
  }{2017}]{wang2017orientation}
Wang, Z.; Tang, L.; Liu, X.; et~al.
\newblock 2017.
\newblock Orientation invariant feature embedding and spatial temporal
  regularization for vehicle re-identification.
\newblock In {\em ICCV},  379--387.

\bibitem[\protect\citeauthoryear{Wang \bgroup et al\mbox.\egroup
  }{2018a}]{wang2018mancs}
Wang, C.; Zhang, Q.; Huang, C.; et~al.
\newblock 2018a.
\newblock Mancs: A multi-task attentional network with curriculum sampling for
  person re-identification.
\newblock In {\em ECCV}.

\bibitem[\protect\citeauthoryear{Wang \bgroup et al\mbox.\egroup
  }{2018b}]{wang2018learning}
Wang, G.; Yuan, Y.; Chen, X.; et~al.
\newblock 2018b.
\newblock Learning discriminative features with multiple granularities for
  person re-identification.
\newblock In {\em ACM MM},  274--282.

\bibitem[\protect\citeauthoryear{Wang \bgroup et al\mbox.\egroup
  }{2019}]{wang2019progressive}
Wang, X.; Hu, J.-F.; Lai, J.-H.; et~al.
\newblock 2019.
\newblock Progressive teacher-student learning for early action prediction.
\newblock In {\em CVPR},  3556--3565.

\bibitem[\protect\citeauthoryear{Yang \bgroup et al\mbox.\egroup
  }{2019}]{yang2019towards}
Yang, W.; Huang, H.; Zhang, Z.; et~al.
\newblock 2019.
\newblock Towards rich feature discovery with class activation maps
  augmentation for person re-identification.
\newblock In {\em CVPR},  1389--1398.

\bibitem[\protect\citeauthoryear{Zhang \bgroup et al\mbox.\egroup
  }{2019}]{zhang2019DSA}
Zhang, Z.; Lan, C.; Zeng, W.; et~al.
\newblock 2019.
\newblock Densely semantically aligned person re-identification.
\newblock In {\em CVPR}.

\bibitem[\protect\citeauthoryear{Zhao \bgroup et al\mbox.\egroup
  }{2017}]{zhao2017spindle}
Zhao, H.; Tian, M.; Sun, S.; et~al.
\newblock 2017.
\newblock Spindle net: Person re-identification with human body region guided
  feature decomposition and fusion.
\newblock In {\em CVPR}.

\bibitem[\protect\citeauthoryear{Zhao, Li, and others}{2017}]{zhao2017deeply}
Zhao, L.; Li, X.; et~al.
\newblock 2017.
\newblock Deeply-learned part-aligned representations for person
  re-identification.
\newblock In {\em ICCV},  3239--3248.

\bibitem[\protect\citeauthoryear{Zheng \bgroup et al\mbox.\egroup
  }{2019a}]{zheng2019re}
Zheng, M.; Karanam, S.; Wu, Z.; et~al.
\newblock 2019a.
\newblock Re-identification with consistent attentive siamese networks.
\newblock In {\em CVPR},  5735--5744.

\bibitem[\protect\citeauthoryear{Zheng \bgroup et al\mbox.\egroup
  }{2019b}]{zheng2019joint}
Zheng, Z.; Yang, X.; Yu, Z.; et~al.
\newblock 2019b.
\newblock Joint discriminative and generative learning for person
  re-identification.
\newblock In {\em CVPR},  2138--2147.

\bibitem[\protect\citeauthoryear{Zheng, Zheng, and
  Yang}{2017}]{zheng2017unlabeled}
Zheng, Z.; Zheng, L.; and Yang, Y.
\newblock 2017.
\newblock Unlabeled samples generated by gan improve the person
  re-identification baseline in vitro.
\newblock In {\em ICCV},  3754--3762.

\bibitem[\protect\citeauthoryear{Zhong \bgroup et al\mbox.\egroup
  }{2017}]{zhong2017random}
Zhong, Z.; Zheng, L.; Kang, G.; Li, S.; and Yang, Y.
\newblock 2017.
\newblock Random erasing data augmentation.
\newblock {\em arXiv preprint arXiv:1708.04896}.

\bibitem[\protect\citeauthoryear{Zhou and Shao}{2017}]{zhou2017cross}
Zhou, Y., and Shao, L.
\newblock 2017.
\newblock Cross-view gan based vehicle generation for re-identification.
\newblock In {\em BMVC}, volume~1,  1--12.

\bibitem[\protect\citeauthoryear{Zhou and Shao}{2018}]{zhou2018aware}
Zhou, Y., and Shao, L.
\newblock 2018.
\newblock Viewpoint-aware attentive multi-view inference for vehicle
  re-identification.
\newblock In {\em CVPR},  6489--6498.

\bibitem[\protect\citeauthoryear{Zhou \bgroup et al\mbox.\egroup
  }{2018}]{zhou2018rocket}
Zhou, G.; Fan, Y.; Cui, R.; et~al.
\newblock 2018.
\newblock Rocket launching: A universal and efficient framework for training
  well-performing light net.
\newblock In {\em AAAI}.

\bibitem[\protect\citeauthoryear{Zhou \bgroup et al\mbox.\egroup
  }{2019}]{zhou2019omni}
Zhou, K.; Yang, Y.; Cavallaro, A.; et~al.
\newblock 2019.
\newblock Omni-scale feature learning for person re-identification.
\newblock {\em ICCV}.

\end{thebibliography}
}

\end{document}